%% file: ctd.tex
\documentclass[letterpaper, 10 pt, conference]{ieeeconf}  

\IEEEoverridecommandlockouts                              

\usepackage{epsfig} 
\usepackage{mathptmx} 
\usepackage{times} 
\usepackage{amsmath} 
\usepackage{amssymb}  
\usepackage{mathrsfs}

\usepackage{graphicx}
\usepackage{subfigure}
\usepackage[T1]{fontenc}
\usepackage{aecompl}
\usepackage{subfigure, subcaption}
\usepackage{multirow}
\usepackage{mathtools}
\usepackage{footnotehyper}

\input{./math_commands.tex}

\title{\LARGE \bf
Stochastic Trajectory Prediction under Unstructured Constraints
}

\author{Hao Ma$^{1,2}$, Zhiqiang Pu$^{1,2}$, Shijie Wang$^{1,2}$, Boyin Liu$^{1,2}$, Huimu Wang$^{3}$, Yanyan Liang$^{4}$, Jianqiang Yi$^{1,2}$
\thanks{This work is supported by the Strategic Priority Research Program of Chinese Academy of Science under Grant No. XDA27030204, the National Natural Science Foundation of China under Grant 62322316, and the National Natural Science Foundation of China under Grant 62073323.}
\thanks{$^{1}$School of Artificial Intelligence, University of Chinese Academy of Sciences, Beijing, China. 
        {\tt\small shijie.wang2022@outlook.com}}%
\thanks{$^{2}$Institute of Automation, Chinese Academy of Sciences, Beijing 100190, China.
        {\tt\small \{mahao2021, zhiqiang.pu, liuboyin2019, jianqiang.yi\}@ia.ac.cn}}%
\thanks{$^{3}$JD.COM, Beijing, China. 
        {\tt\small wanghuimu1@jd.com}}%
\thanks{$^{4}$Macau University of Science and Technology.
        {\tt\small yyliang@must.edu.mo}}%
}

\begin{document}

\maketitle
\thispagestyle{empty}
\pagestyle{empty}

\begin{abstract}

Trajectory prediction facilitates effective planning and decision-making, while constrained trajectory prediction integrates regulation into prediction. Recent advances in constrained trajectory prediction focus on structured constraints by constructing optimization objectives. However, handling unstructured constraints is challenging due to the lack of differentiable formal definitions. To address this, we propose a novel method for constrained trajectory prediction using a conditional generative paradigm, named Controllable Trajectory Diffusion (CTD). The key idea is that any trajectory corresponds to a degree of conformity to a constraint. By quantifying this degree and treating it as a condition, a model can implicitly learn to predict trajectories under unstructured constraints. CTD employs a pre-trained scoring model to predict the degree of conformity (i.e., a score), and uses this score as a condition for a conditional diffusion model to generate trajectories. Experimental results demonstrate that CTD achieves high accuracy on the ETH/UCY and SDD benchmarks. Qualitative analysis confirms that CTD ensures adherence to unstructured constraints and can predict trajectories that satisfy combinatorial constraints.
\end{abstract}

\section{Introduction}

Trajectory prediction plays a pivotal role in ensuring safe and efficient planning in dynamic environments \cite{payeur1995trajectory,hermes2009long,jetchev2009trajectory,choi2021shared,vishnu2023improving}. 
Existing works primarily focus on two key goals. The first is to incorporate richer social and scene information to capture the \textit{interactivity} among agents \cite{gupta2018social,mohamed2020social,yu2020spatio}. As agents interact with each other, their trajectories are partially determined by the games involved in the interactions. The second is to modele the inherent \textit{uncertainties} in trajectories by introducing random variables into models \cite{zhang2019stochastic,salzmann2020trajectron++,gu2022stochastic}, such that stochastic trajectories could be generated through sampling random variables during inference. Researches on \textit{interactivity} and \textit{uncertainty} in trajectory prediction have been extensive, but the role of \textit{constraint} has received comparatively less attention. We argue that explicitly incorporating constraints in trajectory prediction offers two key advantages. First, it ensures that predicted trajectories comply with specific requirements, avoiding unrealistic or infeasible outcomes. Second, by systematically adjusting these constraints, we can generate counterfactual trajectories, thereby enhancing informed planning.

Previous works on constrained trajectory prediction \cite{zhi2021probabilistic, shi20204} mainly focus on structured constraints. Structured constraints could be represented by hand-designed differentiable functions. By including these functions in a loss function, trajectories can be optimized to adhere to the constraints. Although this approach has achieved some successes, it suffers from the limitation of its representational capability, particularly when dealing with unstructured constraints. Unstructured constraints are particularly challengng as they may lack formal definition and could arise from context, observations, or subjective interpretations. This motivates us to develop a method that can perform trajectory prediction under both structured and unstructured constraints.

Conditional generative models \cite{mirza2014conditional, sohn2015learning, ho2020denoising, shoshan2021gan} provide a powerful tool for trajectory prediction under constraints. By mapping a constraint into a vector (typically one-dimensional) and using it as a condition rather than a term in the loss function, we can avoid the requirement for differentiability. This paradigm significantly extend the range of functions to represent constraints. Discrete logic judgments, specific trajectory features, and even neural networks (NNs) could all be used.

In this paper, we introduce a method named controllable trajectory diffusion (CTD) to address trajectory prediction under unstructured constraints.
In CTD, we draw on human-in-the-loop methods \cite{xin2018accelerating,wu2022survey,yuan2022situ}, train a scoring model based on pairwise comparison data \cite{liu2009learning, stiennon2020learning, zhu2023principled} to quantify constraints into scalars. For trajectory prediction, we employ a conditional diffusion model \cite{ho2020denoising}, conditioning on both constraint quantification and historical trajectories. Consequently, CTD can generates trajectories comply with specific constraints during inference.

In summary, our contributions are as follows: \textbf{(i)} we propose solving constrained trajectory prediction by leveraging the degree of conformity to constraints, and \textbf{(ii)} we introduce CTD, which employs a scoring model to effectively represent various unstructured constraints and uses a conditional diffusion model to enforce multiple constraints during inference. Experimental results show that CTD not only achieves excellent performance in terms of Minimum Average Displacement Error (minADE) and Minimum Final Displacement Error (minFDE) on widely used human trajectory datasets, but also adheres effectively to speed and turning constraints. Moreover, CTD demonstrates the capability to satisfy multiple constraints simultaneously.

\section{Related Work}

\textbf{Stochastic trajectory prediction.}
Stochastic trajectory prediction shifts the focus of prediction from a single optimal trajectory to the potential distribution of trajectories. The complete distribution of trajectories contains more information, which can benefit downstream tasks. To effectively model the inherent uncertainty of trajectories, probabilistic models have emerged as a primary approach. \cite{salzmann2020trajectron++} adopt conditional variational autoencoders (CVAE) \cite{sohn2015learning} to predict multi-agent trajectories from heterogeneous data. \cite{zhang2019stochastic} use the variational recurrent neural network (VRNN) \cite{chung2015recurrent} to predict social plausible trajectories in the crowds. The diffusion model \cite{ho2020denoising} recently has demonstrated promising generative capabilities. \cite{gu2022stochastic} employs a conditional diffusion model that encodes historical interactions as a condition. Then, trajectory distributions gradually converge during denoising, ultimately allowing for accurate prediction of trajectory distributions for multi-agents. While these works have focused primarily on interactions and uncertainties. Our work centers on constraints, aiming to incorporate constraints, whether structured or unstructured, into the prediction.

\textbf{Constrained trajectory prediction.} 
While most existing works have implicitly learned constraints from data, they primarily focus on encoding representations of scenes and interactions to fit trajectory distributions. They often lack explicit representations of the constraints themselves, such as physical dynamics, mathematical properties, or social norms. To address this limitation and ensure that predicted trajectories adhere to these known rules, recent works have explored the incorporation of explicit constraints into prediction. This approach has demonstrated the potential to enhance both the accuracy and interpretability of predictions. \cite{shi20204d} predicts the trajectory of flights using a constrained long short-term memory (LSTM) that incorporates physical constraints to improve accuracy. \cite{zhi2021probabilistic} presents a novel framework for probabilistically predicting trajectory while ensuring compliance with environmental limitations (e.g. obstacles) through a hybrid learning and optimization approach. Despite achieving excellent results, these methods hinge upon formulaic differentiable constraints. This reliance on differentiability and explicit formulation restricts their applicability to a subset of constraints. Many real-world constraints are non-differentiable or difficult to formulate mathematically. This motivates our pursuit of more generalized methods capable of predicting trajectories under a broader spectrum of constraints.


\textbf{Controllable generation.} 
In the fields of image and video generation, there is an increasing emphasis on controllable generation, which also focuses on ensuring that the generated content adheres to specified rules. Within video generation, \cite{hao2018controllable} proposes a method for explicit control over synthesized videos via sparse motion trajectories, such as hand-drawn arrows. This enables the generation of subsequent frames aligned with the prescribed trajectory, given an initial frame. In the realm of image generation, \cite{shoshan2021gan} proposes GAN-Control, a framework that empowers precise attribute manipulation in generated images—akin to sculpting digital clay with sliders for age, pose, and expression. \cite{zhang2023adding} further explores the integration of additional conditioning mechanisms, granting users the ability to fine-tune specific details and further personalize the generated imagery. These recent developments in controllable generation inspire us to create a similar framework for constrained trajectory prediction.

\section{Preliminary}
\subsection{Problem Formulation of Trajectory Prediction}

Trajectory broadly refers to a sequence of states. In the context of this paper, we specifically focus on position trajectories. Trajectory prediction is a form of time-series forecasting. It can be expressed as using historical trajectory to predict future trajectory. We take the following notation to denote historical and future trajectories:
\begin{align}
    \vx^{(i)} &= \{\vp_i^{t}, \vp_i^{t+\Delta t}, \cdots, \vp_i^{t+(n-1)\Delta t}\},\\
    \vy^{(i)} &= \{\vp_i^{t+n\Delta t}, \vp_i^{t+(n+1)\Delta t}, \cdots, \vp_i^{t+(n+m-1)\Delta t}\},
\end{align}
where $\vx^{(i)}$ is the history trajectory with a length of $n$ and interval $\Delta t$. $\vy^{(i)}$ is the future trajectory with a length of $m$ and interval $\Delta t$. $\vp_i^t=\{x_i^t, y_i^t\}$ denotes the positions of trajectory $i$ at time step $t$. Following the setting of \cite{gupta2018social}, we set $n=8,m=12$ in experiments. A dataset is constructed by $N$ trajectories: $D=\{\vx^{(i)}, \vy^{(i)}\}_{i=0}^{N-1}$. For simplicity, we can omit the superscripts of $\vx^{(i)}$ and $\vy^{(i)}$, and denote a trajectory using only $\vx$ and $\vy$.

\begin{figure*}[th]
    \centering
    \includegraphics[width=0.85\linewidth]{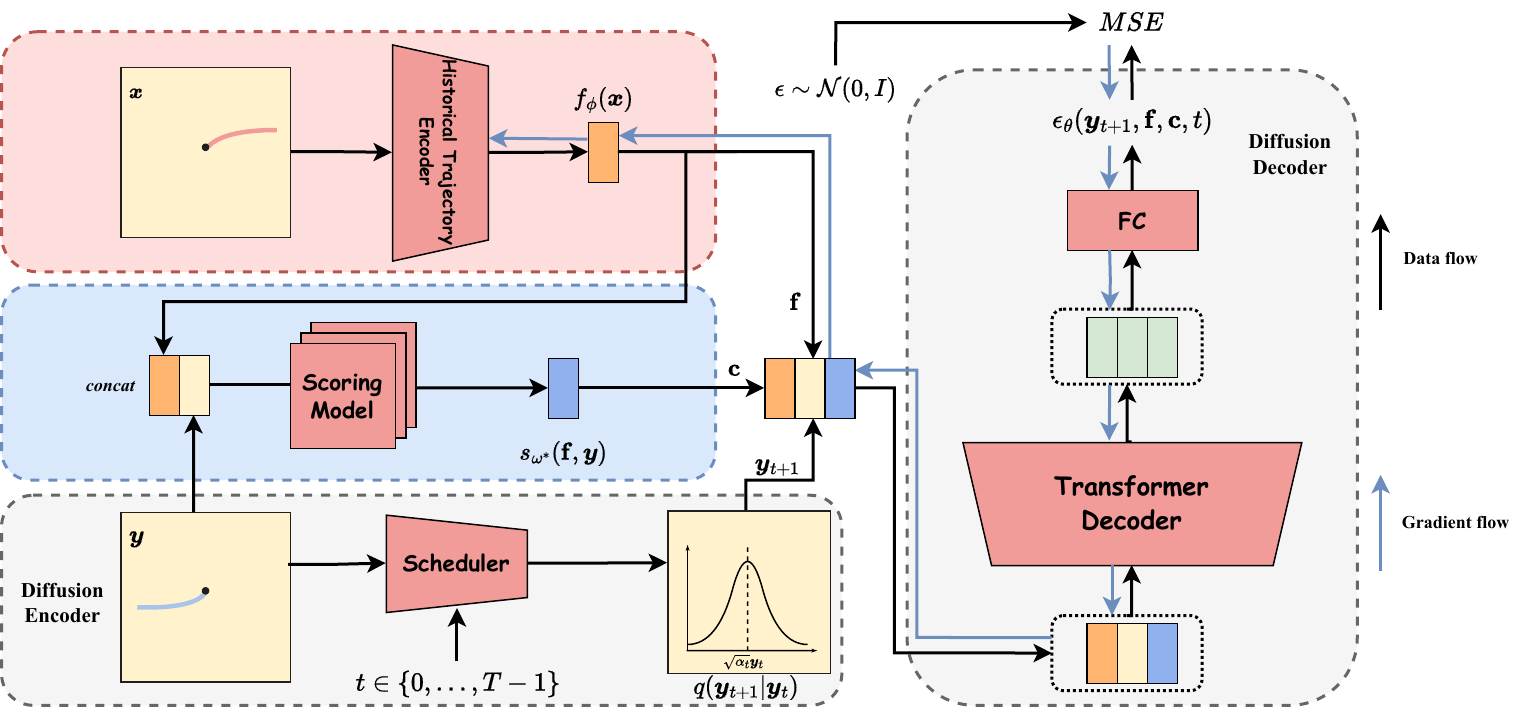}
    \caption{The training pipeline of CTD. The encoder at the top left is responsible for encoding interaction and history information, and the module in the middle left receives $\mathbf{f}$ and $\vy$ and gives the score $c$. The module at the bottom left is the parameterless encoder in diffusion. On the right is the diffusion model's decoder, which is responsible for recovering the original $\vy$ from the noise.}
    \label{fig:pipeline}
\end{figure*}

\begin{figure}[h]
    \centering
    \includegraphics[width=\linewidth]{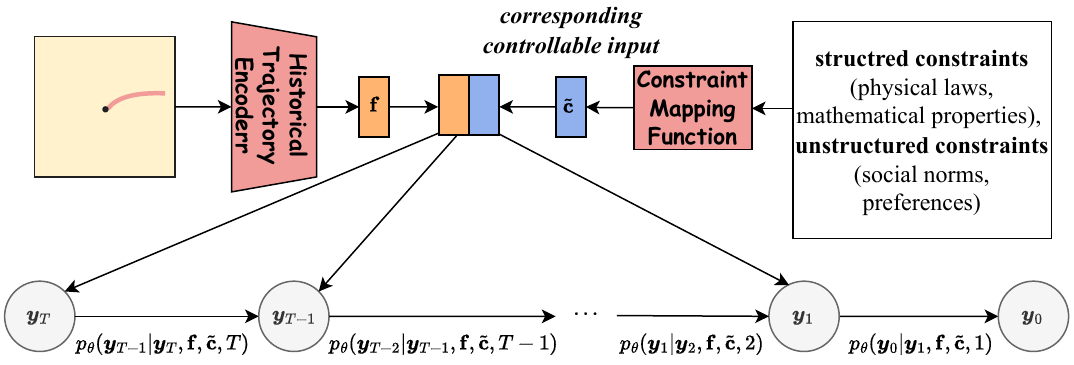}
    \caption{The inference phase of CTD. In the inference process, the predicted trajectory is obtained by iteratively denoising $T$ steps using the decoder $\epsilon_\theta$, conditioned on the encoding $\mathbf{f}$ and the assigned constraint $\tilde{c}$.}
    \label{fig:infer}
\end{figure}

\subsection{Diffusion Probabilistic Models}
A diffusion model \cite{ho2020denoising} is a generative model that can be conceptualized as a Markovian hierarchical variational autoencoder with a fixed encoder (i.e. scheduler). Training involves two stages: noising and denoising. The noising stage gradually adds noise to the original input $\vh_0$ following a predefined scheduler
\begin{equation}
    q(\vh_{t+1}|\vh_t)={N}\left(\vh_{t+1}; \sqrt{\alpha_t}\vh_t, (1-\alpha_t)\mI\right)
\end{equation}
to create a sequence of latent representations, where $\vh_t$ is a vector or a tensor, $t\in [0,1,\dots, T-1]$, $\alpha_t \rightarrow 0$ ($t\rightarrow +\infty$), and $\vh_t \rightarrow {N}\left(0,\mI\right)$ ($t\rightarrow +\infty$). In the denoising stage, a noise $\epsilon$ is sampled from a standard Gaussian distribution ${N}\left(0,\mI\right)$. Then, a denoising process $p(\vh_{t}|\vh_{t+1})$ is trained to reconstruct the original input $\vh_0$ from Gaussian noise. Each denoising step is a Gaussian transition. It gradually denoise the $\vh_T$ by
\begin{equation}
\label{denoising}
    p(\vh_{t-1}|\vh_t) = N(\vh_{t-1};\frac{1}{\sqrt{\alpha_t}}\left(\vh_t-\frac{\beta_t}{\sqrt{1-\overline{\alpha}_t}}\epsilon_\theta(\vh_t,t)\right), \beta_t\mI),
\end{equation}
where $\beta_t=1-\alpha_t$, $\overline{\alpha}_t=\prod_{i=1}^t\alpha_t$, and $\epsilon_\theta(\vh_t,t)$ is the denoising network of diffusion model.



Instead of using evidence lower bound \cite{kingma2013auto} to optimize the diffusion model, a common practice is using the simplified surrogate loss \cite{ho2020denoising}:
\begin{equation}
\label{loss:diffusion}
    {L}(\theta) := \mathbb{E}_{t \sim U\{1, T\},\epsilon \sim {N}\left(0,\mI\right)} \left[ \left\| \epsilon - \epsilon_\theta(\vh_t,t) \right\|^2 \right].
\end{equation}
We refer readers to \cite{luo2022understanding} for detailed derivation.

\subsection{Bradley-Terry-Luce Model}


Bradley-Terry-Luce (BTL) model \cite{bradley1952rank} is a widely use method to train a scoring model from pairwise comparison dataset. 
For a trajectories pair $(\vx,\vy_0,\vy_1)$, BTL model defines the probability that $(\vx,\vy_l)$ is preferred over $(\vx,\vy_{1-l})$ by:
\begin{equation}
    \mathbb{P}(l|\vx, \vy_0, \vy_1) = \frac{exp(s_{\omega^*}(\vx,\vy_l))}{exp(s_{\omega^*}(\vx,\vy_0))+exp(s_{\omega^*}(\vx,\vy_1))},
\end{equation}
where $l\in\{0,1\}$. The key insight of the BTL model is its ability to quantify the conformity to any abstract concept through simple pairwise comparisons. Given a dataset $D_{score}=\{(\vx^{(i)}, \vy^{(i)}_0, \vy^{(i)}_1 , l^{(i)})\}_{i=0}^M$, to train a scoring model, we derive a loss function from maximum likelihood estimation \cite{liu2009learning}:
\begin{equation}
    \label{eq:loss_mle}
    {L}_{mle}(\omega) = -\sum_{i=1}^Mlog \frac{exp(s_\omega(\vx^{(i)},\vy^{(i)}_{l^{(i)}})))} {\sum_{j=0}^{1}exp(s_{\omega}(\vx^{(i)},\vy^{(i)}_j))}.
\end{equation}

\section{Controllable Trajectory Diffusion}


\subsection{Scoring Model} 
Given a historical trajectory $\vx$, a corresponding future trajectory $\vy$, and a constraint, e.g., a conceptual description like `slow speed', our goal is to train a NN to predict the degree $c\in[0,1]$ to which the trajectory pair $\{\vx, \vy\}$ satisfies the constraint. For example, if the trajectory pair $\{\vx, \vy\}$ perfectly meets the `slow speed' constraint, then the predicted value $c$ should be high, otherwise, $c$ should be low. The challenge is how to develop such a scoring model. To address this, we adopt the BTL model for its implementation.


\textbf{Collecting Data.} First, we use a pre-trained trajectory prediction model to generate two distinct future trajectories $\vy_0$ and $\vy_1$ for a history trajectory $\vx$. Noting that achieving exceptional accuracy is not necessary at this stage, generating reasonably accurate trajectories is sufficient. An annotator then labels which of the two trajectory pairs, $(\vx, \vy_0)$ or $(\vx, \vy_1)$, better aligns with the constraint. For instance, if $(\vx, \vy_0)$ is more aligned, the label is set as $l=0$. For unstructured constraints, the annotator may be a human, whereas for structured constraints, it could be an automated algorithm. This process yields a scoring dataset, $D_{score}=\{(\vx^{(i)}, \vy^{(i)}_0, \vy^{(i)}_1 , l^{(i)})\}_{i=0}^M$.

\textbf{Training.} Then, we use $D_{score}$ to train a BTL model $s_{\omega}(\cdot, \cdot)$. During training, we aim to achieve two objectives: 1) maximize the score for the trajectory pair that better aligns with the constraint, $s_\omega(\vx, \vy_l)$, as shown in Eq.~\ref{eq:loss_mle}, and 2) mitigate training instability when the size of $D_{score}$ is small. The instability would make scores to concentrate around 0 or 1, which impairs effective constraint adjustment during inference.

To obtain a more uniform distribution of scores, we adopt a straightforward entropy regularization term as Eq.~\ref{eq:score_loss_penalty}. For each batch, kernel density estimation is employed to estimate the distribution of the scores, denoted as $\hat{p}(\cdot)$. Subsequently, the entropy of this distribution is approximated through sampling. This entropy metric serves as a metric to monitor the extent to which the codomain of $s_{\omega}(\cdot, \cdot)$, is comprehensively covered. By maximizing this entropy penalty, we intentionally induce a broader output distribution while ensuring the preservation of rank order.

The final loss function for the scoring model combines the maximum likelihood estimation loss with the entropy penalty, weighted by the parameter $\lambda$:
\begin{equation}
    {L}_{score}(\omega) = {L}_{mle}(\omega) - \lambda \hat{{H}}(\omega),
    \label{eq:score_loss}
\end{equation}
\begin{equation}
    \hat{{H}}(\omega) = -\sum_{i=1}^K \hat{p}(i/ K)log \hat{p}(i/ K),
    \label{eq:score_loss_penalty}
\end{equation}
where $K$ represents the number of samples used to approximate entropy by evenly sampling the estimated probability density $\hat{p}(\cdot)$.

\subsection{Diffusion over Trajectories} 

As illustrated in Fig.~\ref{fig:pipeline}, CTD leverages conditional diffusion to predict future trajectories under constraints. Three modules are emploied: a historical trajectory encoder, a scoring model, and a conditional diffusion model. The scoring model and conditional diffusion model share the same hisorical trajectory encoder. The data flows through the pipeline in the following order. 1) Historical trajectory encoder: The historical trajectory encoder extracts features from a historical trajectory, denoted as $\mathbf{f}$. 2) Scoring model: The pre-trained scoring model receives features $\mathbf{f}$ and the ground truth future trajectory $\vy$, generating a score $c$ for the entire trajectory $\{\vx, \vy\}$. 3) Conditional diffusion: The historical encoding $\mathbf{f}$ and the constraint score $c$ are jointly used as conditional inputs to a Transformer-based decoder. 

Applying the conditional diffusion model to trajectories reveals fascinating properties. The model demonstrates excellent generation capabilities and enables highly accurate prediction results. Furthermore, the step size $T$ plays a crucial role in controlling the stochasticity of predicted trajectories. A smaller $T$ results in more stochastic predictions \cite{gu2022stochastic}.

When using a diffusion model over trajectory, the noising phase gradually adds Gaussian noise independently to each element of the trajectory vector:
\begin{equation}
    q(\vy_{t} | \vy_{t-1}) = {N}\left(\vy_{t}; \sqrt{\alpha_t} \vy_{t-1}, (1-\alpha_t) \mI \right).
\end{equation}
Each sample in the training set $D$ is assigned a score $c$ by a pre-trained scoring model, indicating its degree of conformity to a constraint. Similar to Eq.~\ref{loss:diffusion}, we use the following loss function to train the denoising network:
\begin{equation}
    {L}_{traj}(\theta) := \mathbb{E}_{t\sim U\{1,N\},\epsilon\sim {N}\left(0,\mI\right)} \left[ \left\| \epsilon - \epsilon_\theta(\vy_t, \mathbf{f}, c, t) \right\|^2 \right].
    \label{eq:loss_ctd}
\end{equation}

\subsection{Prediction with Constraints}
\label{sec:pred_with_con}
As illustrated in Fig.~\ref{fig:infer}, once the denoising function is trained, we could assign a constraint score $\tilde{c}$ and generate trajectory $\vy$ by repeating the denoising function $T$ times according to Eq.~\ref{denoising}:
\begin{equation}
    \vy_{t-1} = \frac{1}{\sqrt{\alpha_t}}(\vy_t-\frac{\beta_t}{\sqrt{1-\bar{\alpha_t}}}\epsilon_\theta(\vy_t, \mathbf{f}, \tilde{c}, t)).
\end{equation}
We continuously adjust the score $\tilde{c}$ within the range $[0, 1]$ to control the generated trajectories, ranging from completely ignoring the constraint to perfectly aligning with it. While this paper primarily focuses on unstructured constraints, the CTD's framework is also compatible with structured constraints, as illustrated in Fig.~\ref{fig:infer}. For structured constraints, factors such as historical speed, acceleration, and goal position can be used as conditions.

Furthermore, CTD could enable trajectories to satisfy multiple constraints by conditioning the diffusion model on multiple constraint scores, i.e., $\epsilon_\theta(\vy_t, \mathbf{f}, {c}_1, {c}_2, \dots, {c}_n, t)$.

\subsection{Implementation Details}
The historical trajectory encoder follows the design in \cite{salzmann2020trajectron++}, consisting of an RNN-based encoder for the target agent's history and an RNN-based edge encoder that aggregates neighboring agents' trajectories. The final feature dimension is 1024 + 512 = 1536.

The scoring model is a MLP with hidden layer sizes of [512, 128, 64], with input and output dimensions of 1536 and 1, respectively. Each layer, except the last, is followed by a LeakyReLU activation, while the final layer uses a Sigmoid function to map the output to the range (0,1).

For the diffusion model, we set the step size to $T=100$. The denoising network, $\epsilon_\theta$, is a TransformerConcatLinear\footnote{https://github.com/Gutianpei/MID/blob/main/models/diffusion.py}, which integrates a transformer architecture with concatenation and squash linear layers.

\section{Experiments}

In this section, we evaluate the performance of CTD on three widely-used datasets: the Stanford Drone Dataset \cite{robicquet2016learning}, UCY dataset \cite{lerner2007crowds}, and ETH dataset \cite{pellegrini2010improving}. We conduct a comprehensive analysis, combining qualitative and quantitative assessments, to address the following questions:

$Q_1$: How is the prediction accuracy of CTD?

$Q_2$: Can CTD effectively satisfy unstructured constraints?

$Q_3$: Can CTD satisfy multiple unstructured constraints at once?
\begin{table*}[th]
    \centering
    \resizebox{0.85\textwidth}{!}{
    \begin{tabular}{l|c|c|c c|c c|c c|c c|c c|c c}
    \hline\hline
        \multirow{2}{*}{Models} & \multirow{2}{*}{Input} & \multirow{2}{*}{Sampling} & \multicolumn{2}{|c|}{ETH} & \multicolumn{2}{|c|}{HOTEL} & \multicolumn{2}{|c|}{UNIV} & \multicolumn{2}{|c|}{ZARA1} & \multicolumn{2}{|c|}{ZARA2} & \multicolumn{2}{|c}{AVG} \\ \cline{4-15}
        ~ & ~ & ~ & minADE & minFDE & minADE & minFDE & minADE & minFDE & minADE & minFDE & minADE & minFDE & minADE & minFDE \\ \hline
        SoPhie \cite{sadeghian2019sophie} & T+I & 20 & 0.70 & 1.43 & 0.76 & 1.67 & 0.54 & 1.24 & 0.30 & 0.63 & 0.38 & 0.78 & 0.54 & 1.15 \\
        CGNS \cite{li2019conditional} & T+I & 20 & 0.62 & 1.40 & 0.70 & 0.93 & 0.48 & 1.22 & 0.32 & 0.59 & 0.35 & 0.71 & 0.49 & 0.97 \\
        Social-BiGAT \cite{kosaraju2019social} & T+I & 20 & 0.69 & 1.29 & 0.49 & 1.01 & 0.55 & 1.32 & 0.30 & 0.62 & 0.36 & 0.75 & 0.48 & 1.00 \\
        MG-GAN \cite{dendorfer2021mg} & T+I & 20 & 0.47 & 0.91 & 0.14 & 0.24 & 0.54 & 1.07 & 0.36 & 0.73 & 0.29 & 0.60 & 0.36 & 0.71 \\
        Y-Net \cite{mangalam2021goals}+TTST & T+I & 10000 & 0.28 & 0.33 & 0.10 & 0.14 & 0.24 & 0.41 & 0.17 & 0.27 & 0.13 & 0.22 & 0.18 & 0.27 \\ \hline
        Social-GAN \cite{gupta2018social} & T & 20 & 0.81 & 1.52 & 0.72 & 1.61 & 0.60 & 1.26 & 0.34 & 0.69 & 0.42 & 0.84 & 0.58 & 1.18 \\ 
        Causal-STGCNN \cite{chen2021human} & T & 20 & 0.64 & 1.00 & 0.38 & 0.45 & 0.49 & 0.81 & 0.34 & 0.53 & 0.32 & 0.49 & 0.43 & 0.66 \\ 
        PECNet \cite{mangalam2020not} & T & 20 & 0.54 & 0.87 & 0.18 & 0.24 & 0.35 & 0.60 & 0.22 & 0.39 & 0.17 & 0.30 & 0.29 & 0.48 \\
        STAR \cite{yu2020spatio} & T & 20 & 0.36 & 0.65 & 0.17 & 0.36 & 0.31 & 0.62 & 0.26 & 0.55 & 0.22 & 0.46 & 0.26 & 0.53 \\ 
        Trajectron++ \cite{salzmann2020trajectron++} & T & 20 & 0.39 & 0.83 & 0.12 & 0.21 & 0.20 & 0.44 & 0.15 & 0.33 & 0.11 & 0.25 & 0.19 & 0.41 \\
        LB-EBM \cite{pang2021trajectory} & T & 20 & 0.30 & 0.52 & 0.13 & 0.20 & 0.27 & 0.52 & 0.20 & 0.37 & 0.15 & 0.29 & 0.21 & 0.38 \\ 
        PCCSNET \cite{sun2021three} & T & 20 & 0.28 & 0.54 & 0.11 & 0.19 & 0.29 & 0.60 & 0.21 & 0.44 & 0.15 & 0.34 & 0.21 & 0.42 \\
        \dag Expert \cite{zhao2021you}& T & 20 & 0.37 & 0.65 & 0.11 & 0.15 & 0.20 & 0.44 & 0.15 & 0.31 & 0.12 & 0.26 & 0.19 & 0.36 \\
        \dag Expert \cite{zhao2021you}+GMM& T & 20$\times$20 & 0.29 & 0.65 & \textbf{0.08}& 0.15 & 0.15 & 0.44 & \textbf{0.11}& 0.31 & \textbf{0.09}& 0.26 & \textbf{0.14}& 0.36 \\ 
        MID \cite{gu2022stochastic} & T & 20 & 0.39 & 0.66 & 0.13 & 0.22 & 0.22 & 0.45 & 0.17 & 0.30 & 0.13 & 0.27 & 0.21 & 0.38 \\ \hline
        CTD-R & T & 20$\times$20 & 0.30 & 0.43 & 0.11 & 0.17 & 0.17 & 0.26 & 0.15 & 0.24 & 0.12 & 0.23 & 0.17 & 0.27 \\ 
        CTD-S & T & 20$\times$20 & \textbf{0.25}& \textbf{0.28} & \textbf{0.08}& \textbf{0.11}& \textbf{0.14}& \textbf{0.17}& 0.14& \textbf{0.19}& 0.10& \textbf{0.13}& \textbf{0.14}& \textbf{0.18}\\ \hline\hline
    \end{tabular}
    }
    \caption{{Quantitative results on the ETH/UCY dataset with minADE/minFDE metric.} T denotes trajectory-only methods and T+I denotes trajectory-and-image methods. \dag \ means results are from  \cite{gu2022stochastic}. CTD-R denotes the CTD model under the constraint of `turn right'. CTD-S denotes the CTD model under the constraint of `slow down'. Lower is better.}
    \label{table1}
\end{table*}

\begin{table}[ht]
    \centering
    \resizebox{0.93\linewidth}{!}{
    \begin{tabular}{l|c|c|c|c}
    \hline\hline
        Models& Input& Sampling& minADE&      minFDE\\ \hline
        CGNS \cite{li2019conditional} & T+I & 20 & 15.60& 28.20\\
        Y-Net \cite{mangalam2021goals}+TTST & T+I & 10000 & 7.85& 11.85\\ \hline
        Social-GAN \cite{gupta2018social} & T & 20 & 27.23& 41.44\\ 
        PECNet \cite{mangalam2020not} & T & 20 & 9.96& 15.88\\
        Trajectron++ \cite{salzmann2020trajectron++} & T & 20 & 8.98& 19.02\\
        LB-EBM \cite{pang2021trajectory} & T & 20 & 8.87& 15.61\\ 
        PCCSNET \cite{sun2021three} & T & 20 & 8.62& 16.16\\
        \dag Expert \cite{zhao2021you}& T & 20 & 10.67& 14.38\\
        \dag Expert \cite{zhao2021you}+GMM& T & 20$\times$20 & 7.65& 14.38\\ 
        MID \cite{gu2022stochastic} & T & 20 & 7.61& 14.30\\ \hline
        CTD-S & T & 20$\times$20 & \textbf{5.47}& \textbf{6.84}\\ \hline\hline
    \end{tabular}
    }
    \caption{{Quantitative results on the SDD with minADE/minFDE metric.} T denotes trajectory-only methods and T+I denotes trajectory-and-image methods. \dag \ means results are from  \cite{gu2022stochastic}. CTD-S denotes the CTD model under the constraint of `slow down'. Lower is better.}
    \label{tab:sdd}
\end{table}

\begin{table}[ht]
    \centering
    \renewcommand\arraystretch{1.2}
    \tabcolsep=0.35cm
    \resizebox{0.8\linewidth}{!}{
    \begin{tabular}{lccccc} \hline \hline
           $N_c$&20&20&  15&  10& 5\\ \hline 
           $N_s$&20&1&  5&  10& 15\\ \hline 
           minADE&5.47&11.24&  8.25&  8.17& 10.27\\
           minFDE&6.84&21.54& 14.41& 14.56&22.01\\ \hline\hline
    \end{tabular}
    }
    \caption{{Ablation results for the combination of $N_c$ and $N_s$.} In addition to 20$\times$20, we tried the remaining four combinations on the SDD dataset to test their minADE and minFDE respectively.}
    \label{tab:abl}
\end{table}

\subsection{Setup}
\textbf{Datasets.} We utilize two widely used public datasets in trajectory prediction, the Stanford Drone Dataset (SDD) \cite{robicquet2016learning} and the ETH/UCY \cite{lerner2007crowds,pellegrini2010improving}.

Stanford Drone Dataset: The Stanford Drone Dataset \cite{robicquet2016learning} is a large-scale collection of aerial videos showcasing diverse agents interacting in complex outdoor environments. Comprised of over 100 aerial videos encompassing more than 20,000 tracked targets, it offers a well-established benchmark for human trajectory prediction.

ETH/UCY: The UCY \cite{lerner2007crowds} and ETH \cite{pellegrini2010improving} dataset contains five unique scenes, denoted as eth, hotel, zara1, zara2, and univ. Each scene features a video capturing pedestrian movement, complemented by detailed trajectory records at each timestamp. These scenes exhibit varying recording frequencies. To ensure data consistency, we resample all trajectories to a frequency of 2.5Hz and transformed them into the world coordinate system. Following the four-in-one-out rule \cite{gupta2018social}, for each scene, we utilize it as a test set while the remaining other scenes as a training set.

\begin{figure}[t]
    \centering
    \subfigure[Prediction results of CTD-R.]{
        \includegraphics[width=0.9\linewidth]{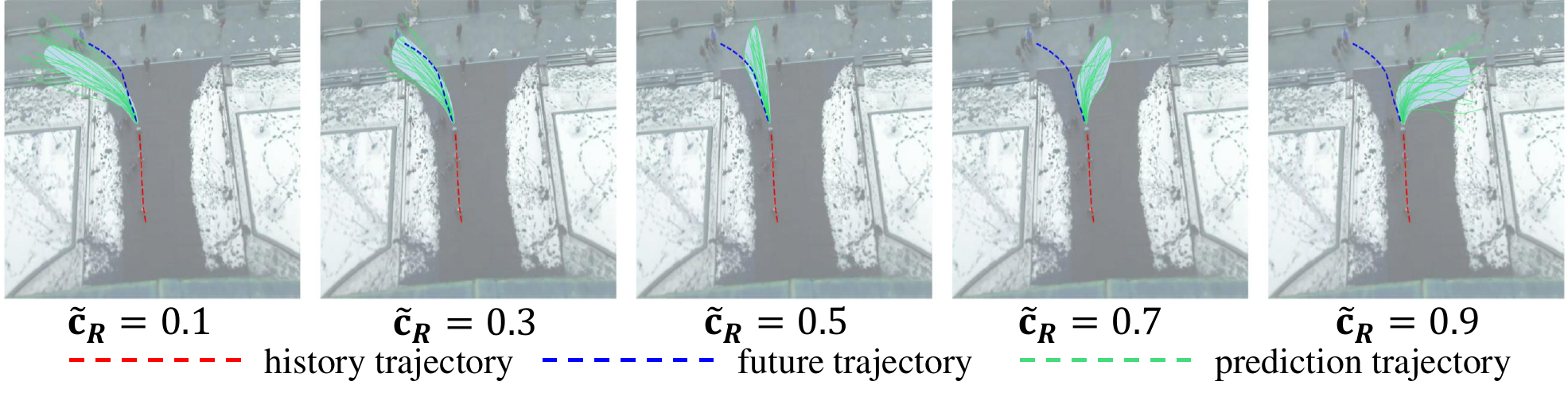}
        \label{fig:exp-preferright}
    }
    \quad
    \subfigure[Prediction results of CTD-S.]{
        \includegraphics[width=0.9\linewidth]{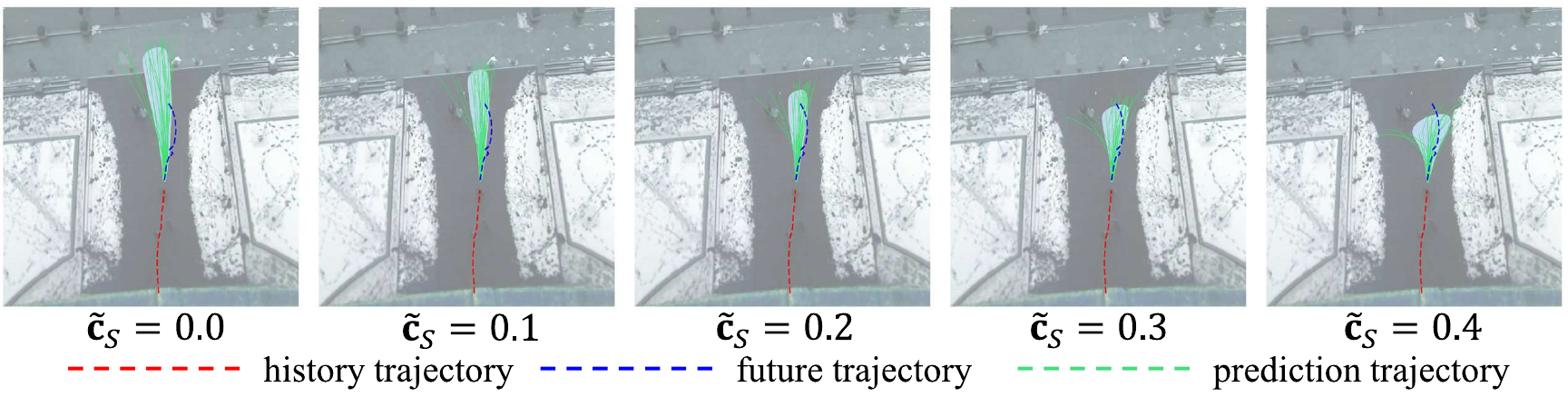}
        \label{fig:exp-preferslow}
    }
     \caption{{Prediction results controlled by different constraints.} At a T-intersection, visual elements demarcate distinct trajectory components: the historical trajectory in red, the true trajectory in blue, the model-sampled trajectory in green, and the distribution of predicted trajectories represented by the blue region. In subfigure (a), $\tilde{c}$ governs the directional bias of future trajectories, with increasing values of $\tilde{c}$ corresponding to a stronger constraint for rightward turns in the predicted trajectory. In subfigure (b), the influence of $\tilde{c}$ remains consistent, but larger values result in a deceleration of the predicted trajectory's speed.}
     \label{fig:single_constrain}
\end{figure}
\textbf{Metrics.}
We use the widely adopted ADE and FDE metrics (Sec.\ref{sec:quant}). ADE measures the overall deviation of the predicted trajectory from the ground truth, while FDE assesses the deviation of the predicted endpoint from the ground truth endpoint. Given the stochastic nature of our method, we apply a Best-of-$N$ strategy to compute the final minADE and minFDE values from $N$ predictions. Furthermore, we conduct qualitative analyses of the predicted trajectories in Sec.\ref{sec:quali}.


\begin{figure*}[t]
    \centering
    \includegraphics[width=0.85\linewidth]{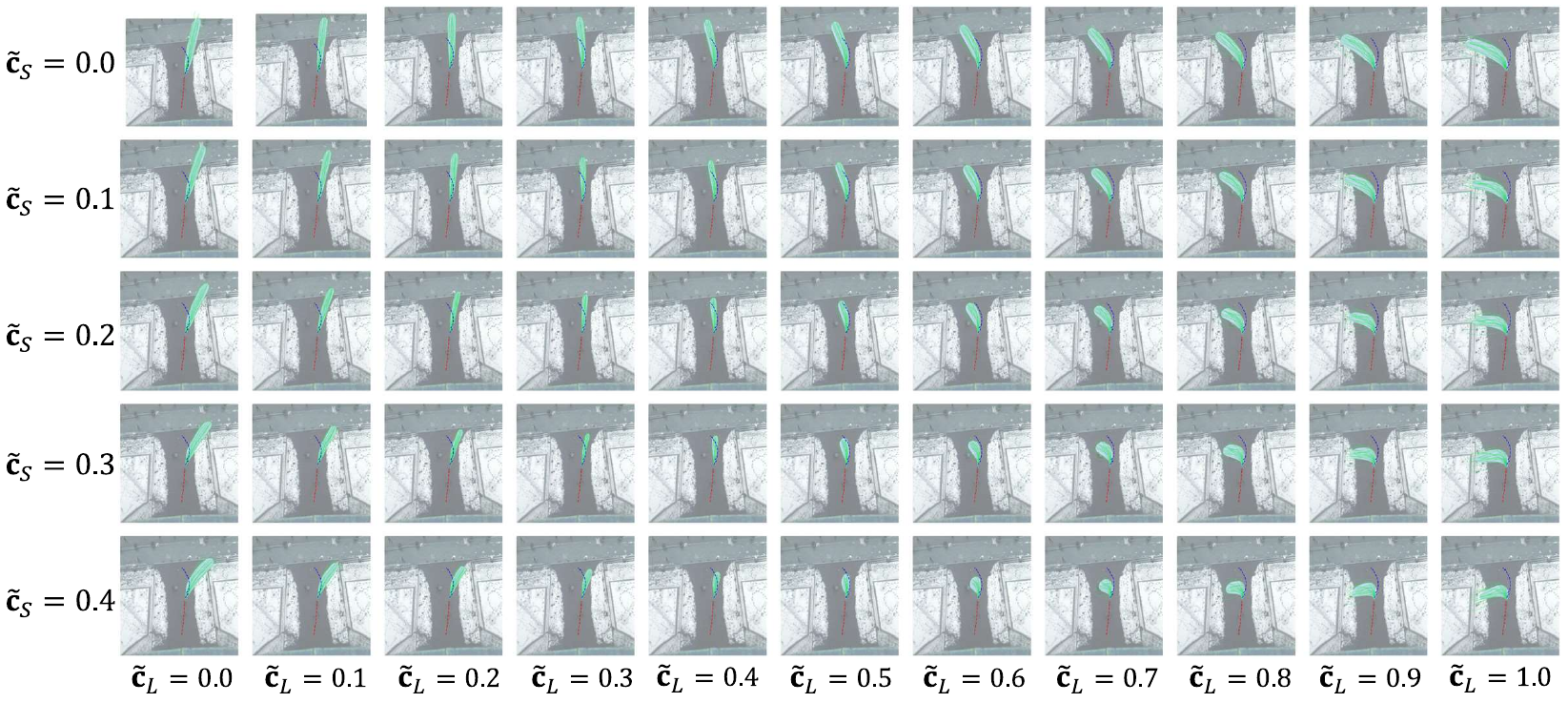}
    \caption{{The combination of two constraints.} The vertical axis represents different speed constraints, with larger controllable input and larger values resulting in slower trajectories. The horizontal axis represents different direction constraints, with larger controllable input and larger values resulting in trajectories that are more inclined towards turning left. During inference, two controllable inputs are concatenated as conditions for the diffusion model. The image shows that CTD can combine multiple constraints.}
    \label{fig:combine_constrians}
\end{figure*}

\subsection{Quantitative comparison to SOTA}
\label{sec:quant}

\textbf{Results}. To answer $Q_1$, we compare the accuracy of CTD against leading state-of-the-art (SOTA) methods in pedestrian trajectory prediction. We evaluate two variants of CTD: one trained under a direction constraint (CTD-R) and another under a speed constraint (CTD-S). For the controllable input $\tilde{c}$, within its codomain of $[0, 1]$, we sample $N_c=20$ values, and for each value, $N_s=20$ trajectories are sampled. The results are shown in Table \ref{table1} and Table \ref{tab:sdd}. Surprisingly, although improving accuracy is not our main concern, CTD-R and CTD-S achieve exceptional minADE and minFDE on both datasets.

\textbf{Analysis.} We assume that the high accuracy of CTD on the benchmark derives from the sampling of $\tilde{c}$. For a trajectory $\{\vx, \vy\}$, if the sampled $\tilde{c}$ is exactly equal to $s_{\omega^*}(f_\Phi(\vx), \vy)$ during inference, the accuracy of the result can be expected to be very high since this is equivalent to goal-conditioned trajectory prediction \cite{mangalam2020not, mangalam2021goals}, where high accuracy is typically achieved with goal information. 

To analyze the impact of the number of samples, we conduct ablation experiments detailed in Table \ref{tab:abl}. The results show that to achieve decent minADE and minFDE scores, both $N_c$ and $N_s$ should be reasonably large, ideally greater than 5. Comparing the parameter sets, we find that increasing both $N_c$ and $N_s$ improves performance. Furthermore, by comparing the parameter sets $N_c=15$, $N_s=5$ and $N_c=5$, $N_s=15$, we find that $N_c$ has a greater impact on results than $N_s$. This confirms our hypothesis: if $\tilde{c}=s_{\omega^*}(f_\Phi(\vx), \vy)$, CTD can achieve high accuracy even with a smaller $N_s$.

\subsection{Qualitative analysis on CTD}
\label{sec:quali}

\textbf{Results.} In this section, we address $Q_2$ and $Q_3$. To evaluate whether CTD can predict trajectories under unstructured constraints, we let annotators label the trajectory data with a paired comparison according to a specific constraint, creating a scoring dataset. Less than 1\% of the training set are labeled. Then, A scoring model is trained on this scoring dataset to quantify trajectories' conformities to the constraint. In the inference stage, we can adjust this score to explicitly control the degree that trajectories conform to the constraint. We trained the model with a constraints of "turn right", "slow down", "turn left + slow down" in ETH dataset, respectively. The visualization results are shown in Fig.~\ref{fig:single_constrain} and Fig.~\ref{fig:combine_constrians}. 

We could see that as $\tilde{c}$ gradually changes from 0 to 1, the generated trajectories change from propensity "turn left" to propensity "turn right" in Fig.~\ref{fig:exp-preferright}, and generated trajectories changes from propensity "high speed" to propensity "slow speed" in Fig.~\ref{fig:exp-preferslow}. That is, the predicted trajectory can be semantically aligned with $\tilde{c}$. Furthermore, Fig.~\ref{fig:combine_constrians} demonstrates that when two constraints are imposed, the generated trajectories can reasonably align with the two constraints, and each constraint can be adjust independently.

\textbf{Analysis.} 
Overall, CTD generates trajectories that are well-aligned with unstructured constraints, with the ability to adjust constraint strength. This leverages the degree of compliance as an additional feature, allowing the neural network to implicitly learn constraint-related knowledge.

Another notable aspect is the distribution of the scoring $c$. Despite entropy regularization aimed at making the distribution of $c$ uniform, the score for the "slow down" constraint is mostly concentrated in the range [0, 0.4], likely due to the limited size of $D_{score}$. This reflects a trade-off between efficiency and effectiveness.

In experiments with combined constraints, CTD exhibited strong performance, highlighting the potential of diffusion models. Conditioning on multiple variables has also shown promise in offline reinforcement learning \cite{ajayconditional}.

\section{Conclusions}
We propose a novel method named CTD for constrained trajectory prediction, which addresses unstructured constraints using a scoring model integrated within a conditional generative framework. By utilizing a scoring model to quantify constraints and employing a conditional diffusion model, CTD generates trajectories that adhere to both individual and combinatorial constraints. Experimental results show that CTD achieves remarkable performance on minADE and minFDE metrics while ensuring compliance with unstructured constraints (e.g., turning, speed), demonstrating its effectiveness and flexibility in constrained trajectory prediction tasks.

\bibliographystyle{IEEEtran}
\bibliography{ctd}

\end{document}

%% file: math_commands.tex
\usepackage{amsmath,amsfonts,bm}

\def\eqref#1{equation~\ref{#1}}

\def\1{\bm{1}}

\def\vh{{\bm{h}}}

\def\vp{{\bm{p}}}

\def\vx{{\bm{x}}}
\def\vy{{\bm{y}}}

\def\mI{{\bm{I}}}

\DeclareMathAlphabet{\mathsfit}{\encodingdefault}{\sfdefault}{m}{sl}
\SetMathAlphabet{\mathsfit}{bold}{\encodingdefault}{\sfdefault}{bx}{n}













%% file: ctd.bbl
\begin{thebibliography}{10}
\providecommand{\url}[1]{#1}
\csname url@samestyle\endcsname
\providecommand{\newblock}{\relax}
\providecommand{\bibinfo}[2]{#2}
\providecommand{\BIBentrySTDinterwordspacing}{\spaceskip=0pt\relax}
\providecommand{\BIBentryALTinterwordstretchfactor}{4}
\providecommand{\BIBentryALTinterwordspacing}{\spaceskip=\fontdimen2\font plus
\BIBentryALTinterwordstretchfactor\fontdimen3\font minus \fontdimen4\font\relax}
\providecommand{\BIBforeignlanguage}[2]{{%
\expandafter\ifx\csname l@#1\endcsname\relax
\typeout{** WARNING: IEEEtran.bst: No hyphenation pattern has been}%
\typeout{** loaded for the language `#1'. Using the pattern for}%
\typeout{** the default language instead.}%
\else
\language=\csname l@#1\endcsname
\fi
#2}}
\providecommand{\BIBdecl}{\relax}
\BIBdecl

\bibitem{payeur1995trajectory}
P.~Payeur, H.~Le-Huy, and C.~M. Gosselin, ``Trajectory prediction for moving objects using artificial neural networks,'' \emph{IEEE Transactions on Industrial Electronics}, vol.~42, no.~2, pp. 147--158, 1995.

\bibitem{hermes2009long}
C.~Hermes, C.~Wohler, K.~Schenk, and F.~Kummert, ``Long-term vehicle motion prediction,'' in \emph{2009 IEEE intelligent vehicles symposium}.\hskip 1em plus 0.5em minus 0.4em\relax IEEE, 2009, pp. 652--657.

\bibitem{jetchev2009trajectory}
N.~Jetchev and M.~Toussaint, ``Trajectory prediction: learning to map situations to robot trajectories,'' in \emph{Proceedings of the 26th annual international conference on machine learning}, 2009, pp. 449--456.

\bibitem{choi2021shared}
C.~Choi, J.~H. Choi, J.~Li, and S.~Malla, ``Shared cross-modal trajectory prediction for autonomous driving,'' in \emph{Proceedings of the IEEE/CVF Conference on Computer Vision and Pattern Recognition}, 2021, pp. 244--253.

\bibitem{vishnu2023improving}
C.~Vishnu, V.~Abhinav, D.~Roy, C.~K. Mohan, and C.~S. Babu, ``Improving multi-agent trajectory prediction using traffic states on interactive driving scenarios,'' \emph{IEEE Robotics and Automation Letters}, vol.~8, no.~5, pp. 2708--2715, 2023.

\bibitem{gupta2018social}
A.~Gupta, J.~Johnson, L.~Fei-Fei, S.~Savarese, and A.~Alahi, ``Social gan: Socially acceptable trajectories with generative adversarial networks,'' in \emph{Proceedings of the IEEE conference on computer vision and pattern recognition}, 2018, pp. 2255--2264.

\bibitem{mohamed2020social}
A.~Mohamed, K.~Qian, M.~Elhoseiny, and C.~Claudel, ``Social-stgcnn: A social spatio-temporal graph convolutional neural network for human trajectory prediction,'' in \emph{Proceedings of the IEEE/CVF conference on computer vision and pattern recognition}, 2020, pp. 14\,424--14\,432.

\bibitem{yu2020spatio}
C.~Yu, X.~Ma, J.~Ren, H.~Zhao, and S.~Yi, ``Spatio-temporal graph transformer networks for pedestrian trajectory prediction,'' in \emph{Computer Vision--ECCV 2020: 16th European Conference, Glasgow, UK, August 23--28, 2020, Proceedings, Part XII 16}.\hskip 1em plus 0.5em minus 0.4em\relax Springer, 2020, pp. 507--523.

\bibitem{zhang2019stochastic}
L.~Zhang, Q.~She, and P.~Guo, ``Stochastic trajectory prediction with social graph network,'' \emph{arXiv preprint arXiv:1907.10233}, 2019.

\bibitem{salzmann2020trajectron++}
T.~Salzmann, B.~Ivanovic, P.~Chakravarty, and M.~Pavone, ``Trajectron++: Dynamically-feasible trajectory forecasting with heterogeneous data,'' in \emph{Computer Vision--ECCV 2020: 16th European Conference, Glasgow, UK, August 23--28, 2020, Proceedings, Part XVIII 16}.\hskip 1em plus 0.5em minus 0.4em\relax Springer, 2020, pp. 683--700.

\bibitem{gu2022stochastic}
T.~Gu, G.~Chen, J.~Li, C.~Lin, Y.~Rao, J.~Zhou, and J.~Lu, ``Stochastic trajectory prediction via motion indeterminacy diffusion,'' in \emph{Proceedings of the IEEE/CVF Conference on Computer Vision and Pattern Recognition}, 2022, pp. 17\,113--17\,122.

\bibitem{zhi2021probabilistic}
W.~Zhi, L.~Ott, and F.~Ramos, ``Probabilistic trajectory prediction with structural constraints,'' in \emph{2021 IEEE/RSJ International Conference on Intelligent Robots and Systems (IROS)}.\hskip 1em plus 0.5em minus 0.4em\relax IEEE, 2021, pp. 9849--9856.

\bibitem{shi20204}
Z.~Shi, M.~Xu, and Q.~Pan, ``4-d flight trajectory prediction with constrained lstm network,'' \emph{IEEE transactions on intelligent transportation systems}, vol.~22, no.~11, pp. 7242--7255, 2020.

\bibitem{mirza2014conditional}
M.~Mirza and S.~Osindero, ``Conditional generative adversarial nets,'' \emph{arXiv preprint arXiv:1411.1784}, 2014.

\bibitem{sohn2015learning}
K.~Sohn, H.~Lee, and X.~Yan, ``Learning structured output representation using deep conditional generative models,'' \emph{Advances in neural information processing systems}, vol.~28, 2015.

\bibitem{ho2020denoising}
J.~Ho, A.~Jain, and P.~Abbeel, ``Denoising diffusion probabilistic models,'' \emph{Advances in neural information processing systems}, vol.~33, pp. 6840--6851, 2020.

\bibitem{shoshan2021gan}
A.~Shoshan, N.~Bhonker, I.~Kviatkovsky, and G.~Medioni, ``Gan-control: Explicitly controllable gans,'' in \emph{Proceedings of the IEEE/CVF international conference on computer vision}, 2021, pp. 14\,083--14\,093.

\bibitem{xin2018accelerating}
D.~Xin, L.~Ma, J.~Liu, S.~Macke, S.~Song, and A.~Parameswaran, ``Accelerating human-in-the-loop machine learning: Challenges and opportunities,'' in \emph{Proceedings of the second workshop on data management for end-to-end machine learning}, 2018, pp. 1--4.

\bibitem{wu2022survey}
X.~Wu, L.~Xiao, Y.~Sun, J.~Zhang, T.~Ma, and L.~He, ``A survey of human-in-the-loop for machine learning,'' \emph{Future Generation Computer Systems}, vol. 135, pp. 364--381, 2022.

\bibitem{yuan2022situ}
L.~Yuan, X.~Gao, Z.~Zheng, M.~Edmonds, Y.~N. Wu, F.~Rossano, H.~Lu, Y.~Zhu, and S.-C. Zhu, ``In situ bidirectional human-robot value alignment,'' \emph{Science robotics}, vol.~7, no.~68, p. eabm4183, 2022.

\bibitem{liu2009learning}
T.-Y. Liu \emph{et~al.}, ``Learning to rank for information retrieval,'' \emph{Foundations and Trends{\textregistered} in Information Retrieval}, vol.~3, no.~3, pp. 225--331, 2009.

\bibitem{stiennon2020learning}
N.~Stiennon, L.~Ouyang, J.~Wu, D.~Ziegler, R.~Lowe, C.~Voss, A.~Radford, D.~Amodei, and P.~F. Christiano, ``Learning to summarize with human feedback,'' \emph{Advances in Neural Information Processing Systems}, vol.~33, pp. 3008--3021, 2020.

\bibitem{zhu2023principled}
B.~Zhu, J.~Jiao, and M.~I. Jordan, ``Principled reinforcement learning with human feedback from pairwise or $ k $-wise comparisons,'' \emph{arXiv preprint arXiv:2301.11270}, 2023.

\bibitem{chung2015recurrent}
J.~Chung, K.~Kastner, L.~Dinh, K.~Goel, A.~C. Courville, and Y.~Bengio, ``A recurrent latent variable model for sequential data,'' \emph{Advances in neural information processing systems}, vol.~28, 2015.

\bibitem{shi20204d}
Z.~Shi, M.~Xu, and Q.~Pan, ``4-d flight trajectory prediction with constrained lstm network,'' \emph{IEEE transactions on intelligent transportation systems}, vol.~22, no.~11, pp. 7242--7255, 2020.

\bibitem{hao2018controllable}
Z.~Hao, X.~Huang, and S.~Belongie, ``Controllable video generation with sparse trajectories,'' in \emph{Proceedings of the IEEE Conference on Computer Vision and Pattern Recognition}, 2018, pp. 7854--7863.

\bibitem{zhang2023adding}
L.~Zhang, A.~Rao, and M.~Agrawala, ``Adding conditional control to text-to-image diffusion models,'' in \emph{Proceedings of the IEEE/CVF International Conference on Computer Vision}, 2023, pp. 3836--3847.

\bibitem{kingma2013auto}
D.~P. Kingma and M.~Welling, ``Auto-encoding variational bayes,'' \emph{arXiv preprint arXiv:1312.6114}, 2013.

\bibitem{luo2022understanding}
C.~Luo, ``Understanding diffusion models: A unified perspective,'' \emph{arXiv preprint arXiv:2208.11970}, 2022.

\bibitem{bradley1952rank}
R.~A. Bradley and M.~E. Terry, ``Rank analysis of incomplete block designs: I. the method of paired comparisons,'' \emph{Biometrika}, vol.~39, no. 3/4, pp. 324--345, 1952.

\bibitem{robicquet2016learning}
A.~Robicquet, A.~Sadeghian, A.~Alahi, and S.~Savarese, ``Learning social etiquette: Human trajectory understanding in crowded scenes,'' in \emph{Computer Vision--ECCV 2016: 14th European Conference, Amsterdam, The Netherlands, October 11-14, 2016, Proceedings, Part VIII 14}.\hskip 1em plus 0.5em minus 0.4em\relax Springer, 2016, pp. 549--565.

\bibitem{lerner2007crowds}
A.~Lerner, Y.~Chrysanthou, and D.~Lischinski, ``Crowds by example,'' in \emph{Computer graphics forum}, vol.~26, no.~3.\hskip 1em plus 0.5em minus 0.4em\relax Wiley Online Library, 2007, pp. 655--664.

\bibitem{pellegrini2010improving}
S.~Pellegrini, A.~Ess, and L.~Van~Gool, ``Improving data association by joint modeling of pedestrian trajectories and groupings,'' in \emph{Computer Vision--ECCV 2010: 11th European Conference on Computer Vision, Heraklion, Crete, Greece, September 5-11, 2010, Proceedings, Part I 11}.\hskip 1em plus 0.5em minus 0.4em\relax Springer, 2010, pp. 452--465.

\bibitem{sadeghian2019sophie}
A.~Sadeghian, V.~Kosaraju, A.~Sadeghian, N.~Hirose, H.~Rezatofighi, and S.~Savarese, ``Sophie: An attentive gan for predicting paths compliant to social and physical constraints,'' in \emph{Proceedings of the IEEE/CVF conference on computer vision and pattern recognition}, 2019, pp. 1349--1358.

\bibitem{li2019conditional}
J.~Li, H.~Ma, and M.~Tomizuka, ``Conditional generative neural system for probabilistic trajectory prediction,'' in \emph{2019 IEEE/RSJ International Conference on Intelligent Robots and Systems (IROS)}.\hskip 1em plus 0.5em minus 0.4em\relax IEEE, 2019, pp. 6150--6156.

\bibitem{kosaraju2019social}
V.~Kosaraju, A.~Sadeghian, R.~Mart{\'\i}n-Mart{\'\i}n, I.~Reid, H.~Rezatofighi, and S.~Savarese, ``Social-bigat: Multimodal trajectory forecasting using bicycle-gan and graph attention networks,'' \emph{Advances in Neural Information Processing Systems}, vol.~32, 2019.

\bibitem{dendorfer2021mg}
P.~Dendorfer, S.~Elflein, and L.~Leal-Taix{\'e}, ``Mg-gan: A multi-generator model preventing out-of-distribution samples in pedestrian trajectory prediction,'' in \emph{Proceedings of the IEEE/CVF International Conference on Computer Vision}, 2021, pp. 13\,158--13\,167.

\bibitem{mangalam2021goals}
K.~Mangalam, Y.~An, H.~Girase, and J.~Malik, ``From goals, waypoints \& paths to long term human trajectory forecasting,'' in \emph{Proceedings of the IEEE/CVF International Conference on Computer Vision}, 2021, pp. 15\,233--15\,242.

\bibitem{chen2021human}
G.~Chen, J.~Li, J.~Lu, and J.~Zhou, ``Human trajectory prediction via counterfactual analysis,'' in \emph{Proceedings of the IEEE/CVF International Conference on Computer Vision}, 2021, pp. 9824--9833.

\bibitem{mangalam2020not}
K.~Mangalam, H.~Girase, S.~Agarwal, K.-H. Lee, E.~Adeli, J.~Malik, and A.~Gaidon, ``It is not the journey but the destination: Endpoint conditioned trajectory prediction,'' in \emph{Computer Vision--ECCV 2020: 16th European Conference, Glasgow, UK, August 23--28, 2020, Proceedings, Part II 16}.\hskip 1em plus 0.5em minus 0.4em\relax Springer, 2020, pp. 759--776.

\bibitem{pang2021trajectory}
B.~Pang, T.~Zhao, X.~Xie, and Y.~N. Wu, ``Trajectory prediction with latent belief energy-based model,'' in \emph{Proceedings of the IEEE/CVF Conference on Computer Vision and Pattern Recognition}, 2021, pp. 11\,814--11\,824.

\bibitem{sun2021three}
J.~Sun, Y.~Li, H.-S. Fang, and C.~Lu, ``Three steps to multimodal trajectory prediction: Modality clustering, classification and synthesis,'' in \emph{Proceedings of the IEEE/CVF International Conference on Computer Vision}, 2021, pp. 13\,250--13\,259.

\bibitem{zhao2021you}
H.~Zhao and R.~P. Wildes, ``Where are you heading? dynamic trajectory prediction with expert goal examples,'' in \emph{Proceedings of the IEEE/CVF International Conference on Computer Vision}, 2021, pp. 7629--7638.

\bibitem{ajayconditional}
A.~Ajay, Y.~Du, A.~Gupta, J.~B. Tenenbaum, T.~S. Jaakkola, and P.~Agrawal, ``Is conditional generative modeling all you need for decision making?'' in \emph{The Eleventh International Conference on Learning Representations}, 2022.

\end{thebibliography}
